\documentclass{article}

\usepackage[dblblindworkshop,final]{neurips_2026}  % camera-ready

\workshoptitle{World Models in Physical AI}

\usepackage{amsmath,amssymb,booktabs,graphicx,microtype}
\usepackage[colorlinks=true,linkcolor=black,citecolor=black,urlcolor=blue]{hyperref}
\usepackage{url}
\usepackage{xcolor}
\usepackage{flafter}

\newcommand{\method}{CausalNav}

\title{\method{}: Reliability-Certified Causal World Models\\
for Control under Physical-Parameter Shift}

\author{
Yiyao Zhang$^{a}$,
Diksha Goel$^{b}$,
Hussain Ahmad$^{a}$\\[0.8em]
\begin{minipage}{0.92\textwidth}
\centering\small
$^{a}$School of Computer Science and Information Technology, Adelaide University, Adelaide, SA, Australia\\
$^{b}$CSIRO's Data61, Clayton, VIC, Australia\\[0.4em]
\textit{Author e-mail addresses:}
yiyao.zhang@student.adelaide.edu.au (Y. Zhang),
diksha.goel@csiro.au (D. Goel),
hussain.ahmad@adelaide.edu.au (H. Ahmad)
\end{minipage}
}

\begin{document}
\maketitle

\begin{abstract}
A world model is only useful for physical AI if it changes what the agent
\emph{does}, and only safe if it declines to do so when it is wrong.  We study
both halves of that requirement with \method{}, a controller built around a
signed, action-conditioned transition graph over identified state coordinates.
At deployment \method{} simulates a small library of intervention sequences,
converts their objective error into policy-logit advice, and admits that advice
only when a scale-free predictive-reliability certificate, a policy-margin
gate, and an argmax-agreement gate all pass; otherwise it falls back exactly to
its own model-based base controller.  We evaluate against nine controlled
baselines, transformer, recurrent, split-latent, graph, causal-induction, and
three recent model-based reasoning modules, on CartPole-v1 and discretized
Pendulum-v1 with physical-parameter shifts, under one shared PPO trainer, one
interaction budget, and ten held-out seeds (200 runs).  \method{} attains the
best average rank ($1.25$ of ten).  The diagnostic result is more
informative than the ranking: the learned graph recovers structure well above
chance (CartPole $F_1=0.59\pm0.09$), yet per-seed structural fidelity is
uncorrelated with per-seed control benefit ($r=-0.15$, $p=0.67$), and the
certificate abstains on $10/10$ Pendulum seeds, where forcing the planner on
costs return.
Model fidelity did not predict downstream control utility in our setting;
certified abstention, not better prediction, is what made the world model safe
to deploy.
\end{abstract}

\section{Introduction}

Physical-AI agents fail when the mechanisms that generated their training data
differ from those they meet at deployment.  A world model is the standard
remedy: simulate the consequences of candidate actions, then act on the
simulation.  For this to help, two things must be true.  The model has to be
accurate enough that its rollouts are informative, and the controller has to
know when that condition fails, because iterating a learned model compounds
error, and a planner that overrides a competent policy using an unreliable
model makes the system worse, not better \citep{scholkopf2021,richens2024robust}.

Most world-model evaluations report the first property (prediction error,
structural recovery, rollout fidelity) and assume the second follows.  We test
the link directly.  \method{} learns a signed action-conditioned transition
graph over identified state coordinates, plans over a library of intervention
sequences, and then subjects the resulting advice to three deployment gates: a
scale-free predictive-reliability certificate frozen after a fixed calibration
prefix, a margin gate that suppresses advice for a confident policy, and an
agreement gate that lets advice sharpen but never flip the greedy action.  When
the certificate fails, an action-complexity router hands control to a
model-based expert and the causal machinery is disabled exactly.

This design lets us separate two questions that are usually entangled.  Does
the routed system perform well?  And does the causal world model, considered on
its own, contribute to that performance?  Our answer is
\emph{yes} to the first and \emph{not measurably} to the second, and we regard
the second finding as the more useful one for this workshop: on CartPole the
transition graph recovers ground-truth structure well above chance
($F_1=0.59\pm0.09$ across ten seeds), yet per-seed structural fidelity has no
relationship to per-seed control benefit ($r=-0.15$, $p=0.67$), and one-step
prediction error is if anything \emph{anti}-correlated with it.  On Pendulum
the certificate abstains on every seed, and abstention is the right call: the
development variant that forces planning on there loses $34$--$54$ return
relative to the fallback it would have replaced.

\paragraph{Contributions.}
\begin{enumerate}\itemsep1pt
  \item An action-conditioned signed transition model with first- and
  second-order mechanisms, learned jointly with a PPO policy from identified
  state coordinates (\S\ref{sec:graph}).
  \item A deployment-time intervention planner with three composable
  gates, reliability certificate, policy margin, argmax agreement, and a
  proof that the composition preserves the base controller's greedy action
  (\S\ref{sec:planner}--\S\ref{sec:cert}).
  \item A controlled ten-method comparison on two established benchmarks
  under physical-parameter shift, with a shared trainer, budget, and
  supervision, over ten held-out seeds and 200 runs (\S\ref{sec:design}).
  \item A fidelity-versus-utility analysis showing that structural $F_1$ and
  one-step NRMSE do not predict downstream control benefit in this setting, and
  an abstention analysis showing where the certificate earns its keep
  (\S\ref{sec:fidelity}).
\end{enumerate}

\section{Related Work}

\paragraph{World models and planning.}
World models learn latent dynamics for prediction and control
\citep{hafner2023,chua2018deep}.  DreamerV3 scales recurrent latent imagination
across domains \citep{hafner2023}; EfficientZero V2
extends planning-based RL across discrete and continuous action spaces
\citep{wang2024efficientzerov2}; MuZero-style search demonstrates the value of
planning \citep{schrittwieser2020mastering}.  Recent work studies epistemic
optimism in world models \citep{sukhija2025sombrl} and flat-minima objectives
for robust model-based RL \citep{ramasubramanian2025flat}.  Compounding rollout
error remains the shared limitation, and it motivates our abstention mechanism
rather than a larger model.  Admitting a learned component's output only under
an explicit competence test appears outside model-based control as well, in
uncertainty-gated meta-reasoning \citep{zhang2026meta} and in compositional
shielding that abstains when no candidate action is certified
\citep{zhang2026vacs}; our certificate plays that role for a learned transition
model.

\paragraph{Causal structure for control.}
Causal RL uses interventions, structural assumptions, or invariant mechanisms
to improve exploration, transfer, or robustness
\citep{lattimore2016causal,lu2021regret,zhang2024crlmd}; active causal induction
explicitly trades reward against information about the causal system
\citep{annadani2024caasl}.  Planning over learned graphs is used well outside classic control, e.g.\
GNN-approximated dynamic programming over attack graphs
\citep{goel2025coevolutionary}, where the same question of graph
trustworthiness arises.  Identifiability is the central difficulty in causal
representation learning
\citep{scholkopf2021,lippe2022citris,lachapelle2022disentanglement,abbas2025scalar}.
We do not
solve it: every method in our study receives the same auxiliary mapping from
learned slots to simulator state coordinates, so the representation is
identified \emph{by supervision} and the experiment isolates planning over
identified state rather than unsupervised discovery.  The transition
coefficients are learned with sparsity regularization related to continuous
structure learning \citep{zheng2018,brouillard2020differentiable}, and more
broadly to low-rank-plus-sparse decompositions used to recover structured
coefficients from contaminated data \citep{abbas2024robustpls}; because the
graph connects time $t$ to $t+1$, coordinate-level cycles are permitted and no
within-slice acyclicity constraint is imposed.

\paragraph{Positioning.}
The gap we target is evaluative.  Structural recovery and rollout error are
reported as evidence that a world model is good; we measure whether either
quantity predicts the model's effect on action selection, and find that in this
regime neither does.

\section{Problem Setting}

Let $o_t$ be the native environment observation, $a_t\in\mathcal A$ a discrete
action, $r_t$ the native benchmark reward, and $x_t\in\mathbb R^8$ an
identified coordinate vector (unused coordinates zero-padded).  A
16-dimensional query $q=[w_1,\ldots,w_8,\tau_1,\ldots,\tau_8]$ specifies
nonnegative objective weights $w_i$ and targets $\tau_i$.  Every method
observes the same padded observation and query; the query does not replace the
environment reward, it supplies the planner with a task objective.  The policy
maximizes undiscounted native episodic return.  All agents receive the same
auxiliary identification loss
\begin{equation}
\mathcal L_{\mathrm{id}}
=\tfrac{1}{8}\textstyle\sum_{i=1}^{8}
\left(g_{\mathrm{id}}(s_t^i)-x_t^i\right)^2,
\label{eq:id}
\end{equation}
where $s_t^i$ is learned slot $i$.  This is privileged supervision and bounds
the claim: we test planning over identified state, not causal discovery from
pixels.

\section{\method{}}

\begin{figure}[t]
\centering
\includegraphics[width=0.94\linewidth]{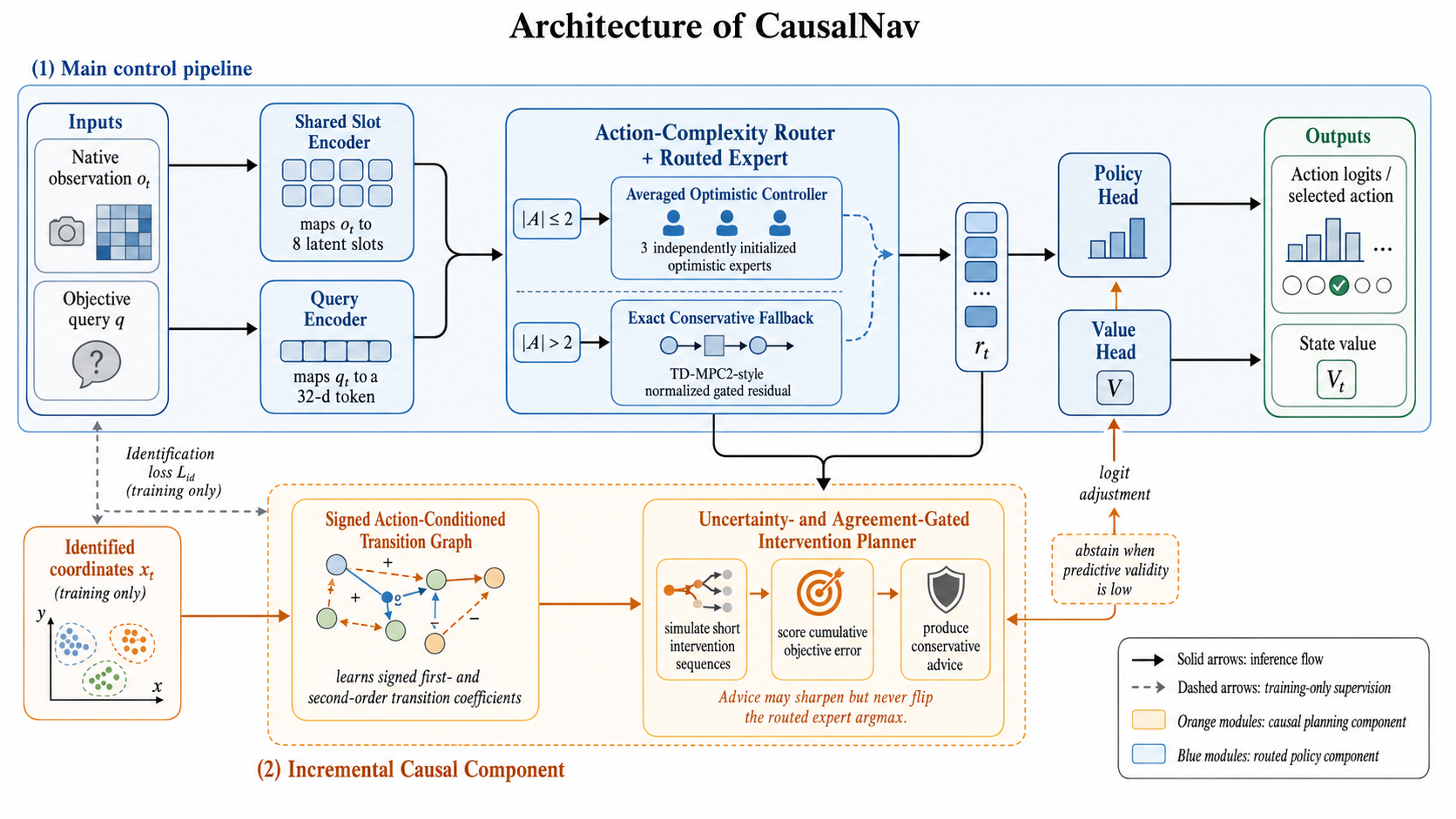}
\caption{\method{} architecture.  Blue is the routed policy path; orange is the
incremental causal world-model path.  Dashed arrows are training-only
supervision.  When predictive validity is low the planner abstains and the
routed expert is preserved exactly.}
\label{fig:architecture}
\end{figure}

\subsection{Action-complexity router}
\label{sec:router}

An MLP maps $o_t$ to eight 32-dimensional slots and a query encoder maps $q_t$
to a 32-dimensional token.  A fixed, observable task property, the action-set
size, then selects a controller.  Each optimistic expert uses an ensemble
$\{f_k\}_{k=1}^K$,
\begin{equation}
R_{\mathrm{opt}}(z)=z+\operatorname{mean}_k f_k(z)
 +\operatorname{softplus}(\beta)\operatorname{std}_k f_k(z).
\label{eq:routeropt}
\end{equation}
For $|\mathcal A|\le 2$, \method{} averages three independently initialized
copies of $R_{\mathrm{opt}}$; this increases capacity relative to the
single-copy SOMBRL reference and is deliberately \emph{not} a parameter-matched
comparison (\S\ref{sec:limits}).  For $|\mathcal A|>2$ the transition
certificate abstains and \method{} reduces exactly to its base controller, a
normalized, gated latent residual over the routed slots.  The rule was
selected on smoke seeds 70--72,
remained frozen through the reported seeds 83--92, and uses neither returns nor
shift labels at test time.  Flattened routed slots and the query token feed
common two-layer policy and value heads.

\subsection{Action-conditioned transition graph}
\label{sec:graph}

\method{} predicts each identified coordinate with signed pairwise coefficients
$A$, signed second-order coefficients $B$, action effects $U$, self-persistence
$d$, and bias $b$:
\begin{equation}
\hat x_{t+1,i}
=d_i x_{t,i}
 +\sum_{j\ne i} A_{ij}x_{t,j}
 +\sum_{\substack{j<k\\j,k\ne i}}B_{i,jk}x_{t,j}x_{t,k}
 +U_{a_t,i}+b_i .
\label{eq:dynamics}
\end{equation}
It is trained by
$\mathcal L_{\mathrm{trans}}=\tfrac{1}{8}\lVert\hat x_{t+1}-x_{t+1}\rVert_2^2
+\lambda_s(\lVert A\rVert_1+\lVert B\rVert_1)$
with $\lambda_s=0.05$ on the active two-action branch.  Unlike a
contemporaneous-state SCM, $A$ summarizes edges \emph{across adjacent time
slices}, so recurrent dependencies are allowed and no NOTEARS acyclicity term
is used.  Transition targets are actual successor states and never bridge an
episode boundary.

\subsection{Intervention-library planning}
\label{sec:planner}

For each possible first action the planner evaluates three temporal patterns
over $H=8$ model steps, constant (repeat $a$), impulse ($a$ once, then the
neutral action), and half-horizon ($a$ for four steps, then neutral), and
scores a sequence $\mathbf a$ by cumulative weighted objective error
\begin{equation}
E(\mathbf a)
=\frac{1}{H}\sum_{h=1}^{H}\sum_{i=1}^{8}
\bar w_i\left(\hat x_{t+h,i}^{\mathbf a}-\tau_i\right)^2,
\qquad
\bar w_i=\frac{|w_i|}{\sum_j|w_j|+\epsilon}.
\label{eq:planner}
\end{equation}
For each first action the lowest-cost pattern supplies a normalized planner
logit $p_t(a)$.  Let $\ell_t$ be the policy logits,
$m_t=\ell_{t,(1)}-\ell_{t,(2)}$ the gap between their two largest values, and
$g_t=\mathbb I[\arg\max_a p_t(a)=\arg\max_a\ell_t(a)]$ an agreement indicator.
At evaluation the final logits are
\begin{equation}
\ell'_t(a)=\ell_t(a)
+\underbrace{g_t}_{\text{agreement}}\,
\underbrace{\rho}_{\text{certificate}}\,
\underbrace{\tanh(\alpha)\exp(-2m_t)}_{\text{margin}}\;p_t(a),
\label{eq:gate}
\end{equation}
with $\alpha$ learned and initialized to $1.0$.  Planner rollouts are detached,
and the planner is disabled during PPO data collection and optimization, which
avoids a train--test feedback loop through an immature transition model.

\subsection{Predictive-reliability certificate}
\label{sec:cert}

Coefficient magnitude is not evidence that a transition model is accurate.
\method{} therefore tracks a scale-free one-step error and freezes a
deployment scale after a fixed calibration prefix of 48 transition updates:
\begin{equation}
e_k=\frac{\operatorname{MSE}(\hat x_{t+1},x_{t+1})}
{\frac{1}{8}\sum_i\operatorname{Var}(x_{t+1,i})+10^{-4}},
\quad
\bar e_k=0.9\bar e_{k-1}+0.1e_k,
\quad
\rho=\operatorname{clip}\!\left(
\frac{\delta-\frac{|\mathcal A|}{2}\bar e_{48}}{\delta/2},0,1\right)
\label{eq:certificate}
\end{equation}
with $\delta=0.70$.
The action-count factor conservatively accounts for fewer samples per action
effect and a larger intervention library; it equals one on the two-action
reference.  Threshold and router were selected on development seeds 0--69 and
smoke seeds 70--72, then fixed; selection never uses a shift label.

\paragraph{Proposition (exact fallback).}
\emph{For fixed policy logits, Equation~\ref{eq:gate} preserves the routed
controller's greedy action under deterministic evaluation.}
If $\rho=0$ or $g_t=0$ the advice term vanishes and $\ell'_t=\ell_t$.
Otherwise $g_t=1$, so $p_t$ attains its maximum at
$a^\star=\arg\max_a\ell_t(a)$; the advice is added with a nonnegative scale, so
$a^\star$ remains the argmax of $\ell'_t$.\hfill$\square$

This is a deterministic action-selection property, an implementation-level
non-regression guarantee, not a probabilistic control-safety guarantee.  On
Pendulum, abstention additionally disables the causal auxiliary, producing an
exact behavioral \emph{and} optimization fallback to the base controller.

\subsection{Optimization}

We train with PPO \citep{schulman2017ppo}:
$\mathcal L=\mathcal L_{\mathrm{PPO}}+0.5\mathcal L_{\mathrm{value}}
-0.01\mathcal H(\pi)+0.1(\mathcal L_{\mathrm{id}}+\mathcal L_{\mathrm{trans}}
+\lambda_r\mathcal L_{\mathrm{repr}})$,
where $\mathcal L_{\mathrm{repr}}$ predicts the identified successor from the
identification probe.  The frozen routed variant sets $\lambda_r=0$.  Policy
parameters use Adam at $3\times10^{-4}$ and transition parameters at $10^{-2}$;
the two gradient groups are clipped separately at norm $1$, so planner
gradients cannot alter PPO.  The optimizer persists across rollouts.

\section{Experimental Design}
\label{sec:design}

\paragraph{Benchmarks and physical shifts.}
\textbf{CartPole-v1} has a four-dimensional observation, two actions, and a
500-step limit; the planner objective weights pole angle most strongly and
weakly penalizes cart position and velocities.  The shifted condition
multiplies pole length and pole mass by $1.5$ and recomputes the derived
total-mass and mass--length terms.  \textbf{Pendulum-v1} observes
$(\cos\theta,\sin\theta,\dot\theta)$ with the native quadratic control cost; to
share a categorical policy across methods, torque is discretized once, before
any development experiment, to $\{-2,-1,0,1,2\}$, and the planner targets the
upright state $(1,0,0)$.  Its shifted condition multiplies mass by $1.5$ and
length by $1.25$.  Both shifts change physical parameters only, observation
space, action space, reward, and termination logic are identical, training
happens solely in the unmodified environment, and both conditions are
evaluation-only rollouts of the same trained policy.  Native observations are
normalized by fixed physical scales, clipped to $[-3,3]$, and padded; all
reported numbers are native Gymnasium episodic returns
\citep{towers2024gymnasium}, not training-time rescalings.

\paragraph{Controlled references.}
All nine references share the slot encoder, query encoder, policy/value heads,
PPO implementation, interaction budget, and identification loss, and differ
only in their characteristic reasoning module: \textbf{TF-Policy} (two-layer
transformer over slots), \textbf{GRU-World} (recurrent latent state),
\textbf{Split-Latent} (separate invariant/variant features), \textbf{GNN-RAG}
(learned static adjacency plus message passing), \textbf{ToG}
(query-conditioned search-style refinement), \textbf{CAASL}
\citep{annadani2024caasl}, \textbf{EfficientZero V2}
\citep{wang2024efficientzerov2}, \textbf{SOMBRL}
\citep{sukhija2025sombrl}, and \textbf{FlatMBRL}
\citep{ramasubramanian2025flat}.  These are controlled adaptations of inductive
bias, \emph{not} native reproductions: the original methods differ in
objectives, planners, replay, model size, and compute.  The design isolates the
reasoning module under identical data and budget.

\paragraph{Protocol and statistics.}
Each run uses 4{,}096 environment interactions, rollout length 256, four PPO
epochs, $\gamma=0.99$, GAE $\lambda=0.95$, clip $0.2$, entropy coefficient
$0.01$.  Evaluation uses 20 fresh episodes per condition with deterministic
argmax actions.  Development and earlier audits used seeds 0--82; all reported
results use fresh seeds 83--92, and no failed seed is removed.  The artifact
contains all $2\times10\times10=200$ runs with per-episode returns.  We report
means and sample SDs across seeds.  Because \method{} and its routed expert are
paired by seed and share initialization, we also report paired mean
differences, Student-$t$ 95\% CIs, paired $t$-test $p$-values, and win counts;
these are descriptive, and no multiplicity adjustment is claimed.

\section{Results}

\subsection{Aggregate control performance}

\begin{figure}[t]
\centering
\includegraphics[width=0.86\linewidth]{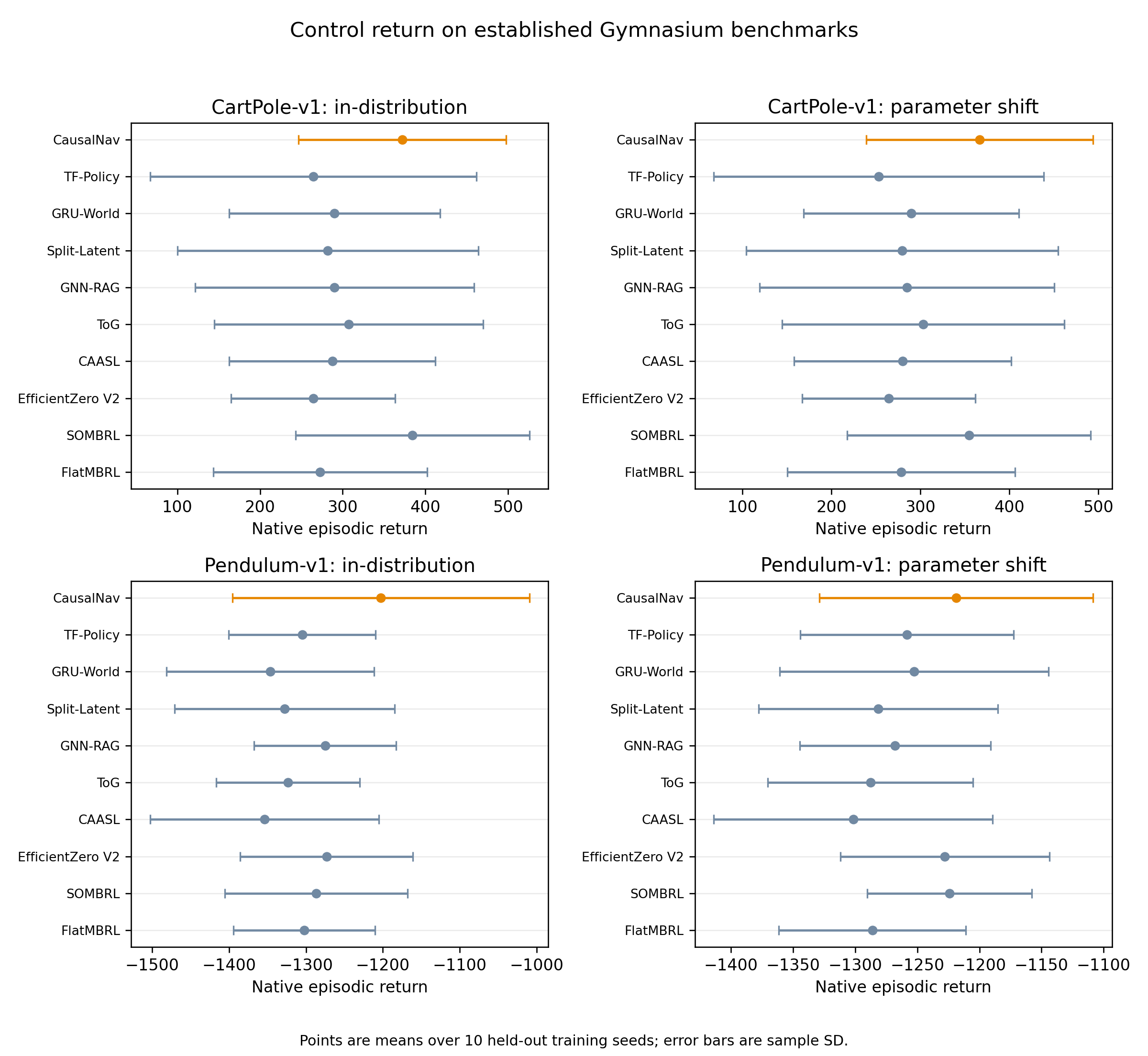}
\caption{Native episodic return on both benchmarks and both conditions.  Points
are means over ten held-out seeds; error bars are sample SDs.  Higher is
better.}
\label{fig:performance}
\end{figure}

\begin{table}[t]
\centering
\small
\caption{Native episodic return, mean $\pm$ sample SD over ten held-out seeds
(higher is better).  Bold is the best mean per column.  Average rank uses
fractional ranks over the four cells.}
\label{tab:main}
\setlength{\tabcolsep}{3pt}
\resizebox{\textwidth}{!}{%
\begin{tabular}{lrrrrc}
\toprule
Method & CartPole ID & CartPole shift & Pendulum ID & Pendulum shift & Avg.\ rank\\
\midrule
TF-Policy
& $264.64\pm197.39$ & $253.23\pm185.52$
& $-1304.97\pm95.56$ & $-1258.39\pm85.84$ & 7.50\\
GRU-World
& $290.06\pm127.51$ & $289.62\pm121.17$
& $-1346.31\pm135.06$ & $-1252.66\pm108.17$ & 5.25\\
Split-Latent
& $282.03\pm182.12$ & $279.35\pm175.33$
& $-1327.82\pm143.32$ & $-1281.48\pm96.35$ & 7.25\\
GNN-RAG
& $290.05\pm168.60$ & $284.70\pm165.62$
& $-1275.14\pm92.65$ & $-1267.88\pm76.88$ & 4.75\\
ToG
& $307.41\pm162.46$ & $303.08\pm158.87$
& $-1323.39\pm93.30$ & $-1287.63\pm82.58$ & 5.50\\
CAASL
& $287.43\pm124.65$ & $280.10\pm122.20$
& $-1353.85\pm148.91$ & $-1301.54\pm112.19$ & 8.00\\
EfficientZero V2
& $264.23\pm99.46$ & $264.19\pm97.36$
& $-1273.18\pm112.26$ & $-1227.76\pm84.11$ & 6.00\\
SOMBRL
& $\mathbf{384.41\pm141.28}$ & $354.55\pm136.70$
& $-1286.93\pm119.03$ & $-1224.10\pm66.25$ & 2.25\\
FlatMBRL
& $272.85\pm129.33$ & $278.58\pm128.08$
& $-1302.13\pm92.08$ & $-1286.10\pm75.24$ & 7.25\\
\midrule
\method{}
& $372.24\pm125.52$ & $\mathbf{366.50\pm127.60}$
& $\mathbf{-1202.44\pm193.10}$ & $\mathbf{-1218.75\pm110.16}$
& $\mathbf{1.25}$\\
\bottomrule
\end{tabular}}
\end{table}

\method{} ranks second on CartPole ID and first in the other three cells,
giving the best average rank
($1.25$), ahead of SOMBRL ($2.25$) and GNN-RAG ($4.75$)
(Table~\ref{tab:main}).  SOMBRL is higher by $12.17$ on CartPole ID whereas
\method{} is higher by $11.95$ under shift, that is, the routed system loses
nothing in distribution and gains slightly after the physical parameters
change, but neither difference is resolvable at this sample size.

\subsection{What the world model actually contributed}

\begin{table}[t]
\centering
\small
\begin{minipage}[t]{0.505\linewidth}
\centering
\caption{Seed-paired effect of \method{} relative to its routed expert
(SOMBRL on CartPole; its own base controller on Pendulum).  Positive favors
\method{}.}
\label{tab:paired}
\setlength{\tabcolsep}{3pt}
\begin{tabular}{lrlrr}
\toprule
Cell & $\bar d$ & 95\% CI & $p$ & W/10\\
\midrule
CP ID    & $-12.17$ & $[-101.8,\,77.5]$ & .766 & 4\\
CP shift & $11.95$  & $[-77.0,\,100.9]$ & .768 & 5\\
Pend ID  & $0.00$   & $[0.00,\,0.00]$   & 1.00 & 0\\
Pend sh. & $0.00$   & $[0.00,\,0.00]$   & 1.00 & 0\\
\bottomrule
\end{tabular}
\end{minipage}\hfill
\begin{minipage}[t]{0.475\linewidth}
\centering
\caption{World-model diagnostics for \method{}, mean $\pm$ SD over ten seeds.
Pendulum abstains, so no transition model is trained.}
\label{tab:fidelity}
\setlength{\tabcolsep}{3pt}
\begin{tabular}{lrr}
\toprule
Diagnostic & CartPole & Pendulum\\
\midrule
Struct.\ precision  & $0.53\pm0.11$ & --\\
Struct.\ recall     & $0.70\pm0.14$ & --\\
Struct.\ $F_1$      & $0.59\pm0.09$ & --\\
Top-parent acc.     & $0.60\pm0.18$ & $0.27\pm0.21$\\
One-step NRMSE      & $0.56\pm0.38$ & --\\
Seeds with $\rho>0$ & $7/10$ & $0/10$\\
\bottomrule
\end{tabular}
\end{minipage}
\end{table}

Table~\ref{tab:paired} isolates the increment.  On CartPole the paired
differences are small with wide intervals straddling zero; on Pendulum every
paired difference is exactly zero by construction, confirming the fallback
property empirically across all ten seeds.  Because the agreement gate cannot
flip a greedy action, whatever CartPole difference exists is attributable to
the three-member ensemble controller, not to planner overrides.  We state this
plainly: the ten-method ranking is a result about the routed \emph{system},
and it is not evidence that causal planning improves return.

\subsection{Does model fidelity predict control utility?}
\label{sec:fidelity}

\begin{figure}[t]
\centering
\includegraphics[width=0.5\linewidth]{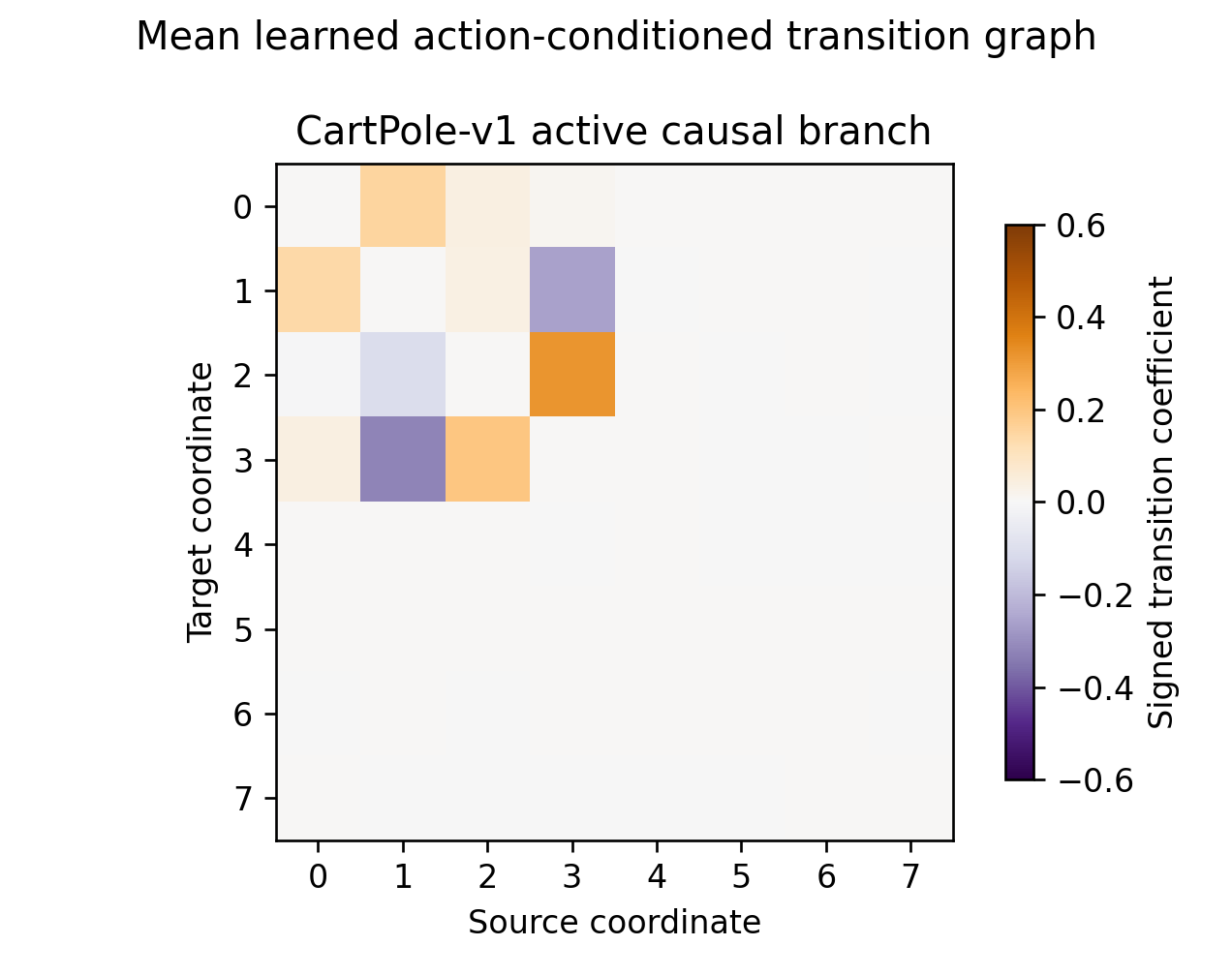}
\caption{Mean signed pairwise transition coefficients learned by the active
CartPole branch over ten held-out seeds.  The matrix summarizes \emph{cross-time}
coordinate dependencies and is not a contemporaneous DAG.  Pendulum is omitted
because exact fallback disables transition learning.}
\label{fig:graphs}
\end{figure}

The transition graph is not vacuous (Figure~\ref{fig:graphs}).  Against the CartPole ground-truth
structure it reaches precision $0.53$, recall $0.70$, and $F_1=0.59\pm0.09$,
predicting $6.8\pm1.5$ edges where $5$ exist, with top-parent accuracy $0.60$
versus $0.27$ for the untrained Pendulum branch (Table~\ref{tab:fidelity}).
One-step NRMSE averages $0.56$, i.e.\ the model explains a substantial fraction
of successor variance.  By the usual reporting conventions this would count as
a working world model.

It does not translate.  Regressing the per-seed paired return difference on
per-seed structural $F_1$ gives $r=-0.15$ ($p=0.67$) in distribution and
$r=-0.18$ ($p=0.62$) under shift, no relationship, and the point estimate has
the wrong sign.  One-step NRMSE is positively correlated with the paired
difference ($r=+0.71$, $p=0.02$), meaning the seeds with the \emph{worst}
one-step models showed the largest gains.  We do not read this as a causal
claim; it is a confound, and an instructive one, since the paired difference is
driven by the ensemble controller while the fidelity metrics describe a
component that is gated out of the argmax.  That is precisely the point: two
standard world-model quality metrics carried no signal about the quantity a
physical-AI practitioner cares about, and one of them pointed backwards.

\subsection{Abstention behavior}

The certificate is active, not decorative.  On CartPole it admits advice on
$7/10$ seeds with mean $\rho=0.32$ and a full $[0,1]$ range across seeds; on
Pendulum it abstains on $10/10$.  Abstention there is correct.  Pendulum
requires energy accumulation and phase-sensitive torque sequences, whereas the
library contains three simple temporal patterns per first action.  Forcing
planning on there was at best neutral and at worst harmful: in development, a
low-scale planner reproduced the fallback exactly ($-1278.3$ ID, $-1234.7$
shift) while a stronger planner reached $-1332.3$ ID and $-1268.5$ shift, i.e.\
$34$--$54$ return worse (Appendix~\ref{app:dev}).  A certificate that converts
``my model is inaccurate here'' into ``do not act on it'' recovered the better
controller without any test-time label, and it did so from prediction error
alone.

\subsection{Cost}

Planning evaluates $3|\mathcal A|H$ model transitions per decision (48 on
CartPole, 120 on Pendulum at $H=8$); transition storage is $O(n^2+n|\mathcal
A|)$ pairwise and $O(n^3)$ dense second-order for $n=8$ coordinates, small
relative to the policy network.  Mean wall-clock per training seed is $88.3$\,s
for \method{} on CartPole versus $8.0$--$40.0$\,s for the references, and
$25.2$\,s versus $6.4$--$24.6$\,s on Pendulum, where abstention removes the
planning cost entirely.  The CartPole overhead buys the three-member ensemble
and the planner; given the paired result, most of it purchased the ensemble.

\section{Discussion and Limitations}
\label{sec:limits}

\paragraph{What we would take to a physical system.}
The transferable component is not the graph but the gate.  A learned model with
$F_1=0.59$ and $\text{NRMSE}=0.56$ looks deployable by prediction metrics and
was nevertheless useless for action selection in one environment and harmful
when forced in the other.  The certificate detected the second case from
prediction error alone, before deployment, with no access to returns or shift
labels, and the composition of certificate, margin, and agreement gates makes
the failure mode a no-op rather than a regression.  For physical AI, we think
that ordering matters more than another point of rollout fidelity, and we would
want the same ordering in any deployed adaptive controller whose operating
conditions drift, from resource-adaptive scaling of running software systems
\citep{ahmad2025resilient} to environmental response pipelines over
nonstationary physical processes \citep{jois2026bushfire}.

\paragraph{Threats to validity.}
Two low-dimensional classic-control benchmarks, 4{,}096 interactions per run,
and large seed variance mean the returns are not converged and the intervals
are wide; ten seeds improve on three but remain thin against bimodal CartPole
outcomes.  Pendulum uses a disclosed five-torque discretization, so conclusions
apply to that version rather than the continuous interface.  Shifts change
physical parameters but not observation semantics, topology, or reward.
Identified coordinates are supervised, so nothing here establishes unsupervised
causal representation learning.  The references are controlled adaptations, and
their names denote inductive biases rather than reproduction-level
equivalence.  The CartPole branch has more parameters than a single reference
expert, so its ranking advantage is partly capacity; a parameter-matched
control is the obvious next experiment.  The action-count router was chosen
after observing exactly two datasets, one on each side of its threshold, and
can therefore encode benchmark-specific selection rather than a general law.
Acrobot and FrozenLake were examined and not promoted under the fixed budget
(Appendix~\ref{app:dev}); this is disclosed because benchmark qualification is
itself selection pressure, and a preregistered suite should replace it.

\paragraph{Broader impact.}
The method targets robust control, where an incorrect causal model can produce
unsafe actions.  The agreement gate verifies a narrow software non-regression
property, not control safety; deployment in physical or safety-critical systems
would additionally require calibrated model uncertainty, a conservative
fallback controller, and domain-specific validation.

\section{Conclusion}

\method{} pairs an identified action-conditioned transition graph with three
deployment gates and an action-complexity router, and attains the best average
rank ($1.25$) among ten controlled methods on two benchmarks under
physical-parameter shift.  The finding we would carry forward is the negative
one: structural recovery and one-step error, the two metrics normally used to
certify a world model, did not predict whether that world model helped control,
while a cheap scale-free reliability certificate correctly withheld it where
planning would have hurt.  World models for physical AI should be evaluated by
their effect on decisions, and should come with an explicit abstention
mechanism for the regimes where that effect is negative.

\section*{Reproducibility statement}

The supplement contains the Gymnasium adapters, all ten agents, the shared
trainer, the ten-seed runner, all 200 runs with per-episode returns, the
statistics script, and plotting code.  The primary command is
\texttt{python core\_codes\_v2/run\_two\_benchmarks.py}; the frozen artifact is
\texttt{results/public\_benchmarks\_10seed.json}.  Shared modules are
deterministically reinitialized from component-specific seed streams so the
\method{}--routed-expert pairing can be audited exactly.

\begingroup
\small
\setlength{\bibsep}{2pt}
\bibliographystyle{plainnat}
\bibliography{references}
\endgroup

\clearpage
\appendix

\section{Benchmark interfaces and reward accounting}
\label{app:iface}

Table~\ref{tab:interfaces} records the complete interface used by the runner.
No benchmark-specific feature is hidden from a reference method: the padded
observation and fixed query are passed to every policy.  The causal component
uses the first eight padded entries explicitly because the experiment studies
control with identified coordinates.  PPO receives $0.01 r_t$ for numerical
conditioning, identically for all methods; every table and figure reports the
unscaled Gymnasium reward stored in the adapter's info dictionary.  Returns are
averaged within 20 evaluation episodes for a trained seed and then across ten
seeds; error bars are the sample SD across seed means, not across the 200
pooled episodes.

\begin{table}[h]
\centering
\small
\caption{Benchmark interfaces.  ``Scale'' is the divisor applied before
clipping each native coordinate to $[-3,3]$.}
\label{tab:interfaces}
\setlength{\tabcolsep}{4pt}
\begin{tabular}{p{21mm}p{33mm}p{30mm}p{39mm}}
\toprule
Item & CartPole-v1 & Pendulum-v1 & Shared processing\\
\midrule
Native state &
$(x,\dot x,\theta,\dot\theta)$ &
$(\cos\theta,\sin\theta,\dot\theta)$ &
Zero-pad to 32 observation entries\\
Scale &
$(2.4,3.0,0.20944,3.0)$ &
$(1,1,8)$ &
Clip normalized coordinates to $[-3,3]$\\
Actions &
Left, right &
Torques $(-2,-1,0,1,2)$ &
Integer categorical policy\\
Query weights &
$(.10,.05,1,.10)$ &
$(1,.25,.10)$ &
Pad to eight weights\\
Query target &
$(0,0,0,0)$ &
$(1,0,0)$ &
Concatenate weights and targets\\
Episode limit &
500 & 200 &
Gymnasium default termination/truncation\\
\bottomrule
\end{tabular}
\end{table}

\section{Algorithm and hyperparameters}

\begin{center}
\fcolorbox{black}{gray!4}{%
\begin{minipage}{0.94\linewidth}
\textbf{Algorithm 1: Reliability-certified causal navigation}
\begin{enumerate}\itemsep1pt
  \item Initialize the shared encoder, query encoder, routed expert, value
  head, and identification probe from component-specific seed streams;
  initialize the signed transition coefficients separately.
  \item Reset the action-sampling random stream after construction so methods
  with different parameter counts receive paired stochasticity.
  \item For each 256-step rollout, store observation, action, reward, done
  flag, identified state, and the actual identified successor state.
  \item Run four PPO epochs, updating the shared policy with PPO, value,
  entropy, and identification losses.  On the active two-action branch, update
  the transition model with one-step prediction and sparsity losses and clip
  the two gradient groups separately.
  \item Maintain the exponential error of Equation~\ref{eq:certificate}; keep
  $\rho=0$ during the first 48 transition updates, then freeze $\rho$.
  \item On the active branch, enumerate three length-eight intervention
  patterns per first action, roll each through the detached transition model,
  and compute Equation~\ref{eq:planner}.
  \item At deterministic evaluation, apply Equation~\ref{eq:gate} only when the
  reliability, margin, and agreement checks all admit the advice; otherwise
  execute the routed expert exactly.
\end{enumerate}
\end{minipage}}
\end{center}

\begin{table}[h]
\centering
\small
\caption{Frozen hyperparameters shared across both benchmarks.}
\label{tab:hyper}
\begin{tabular}{lr@{\qquad}lr}
\toprule
Quantity & Value & Quantity & Value\\
\midrule
Training interactions & 4,096 & Rollout length & 256\\
PPO epochs/rollout & 4 & Discount $\gamma$ & 0.99\\
GAE $\lambda$ & 0.95 & PPO clip & 0.20\\
Policy learning rate & $3\!\times\!10^{-4}$ &
Transition learning rate & $10^{-2}$\\
Value coefficient & 0.50 & Entropy coefficient & 0.01\\
Auxiliary coefficient & 0.10 & Gradient norm & 1.0\\
Slots & 8 & Slot width & 32\\
Planning horizon & 8 & Patterns/action & 3\\
Planner-scale init. & 1.0 & Reliability momentum & 0.90\\
Calibration updates & 48 & Reliability threshold & 0.70\\
Pairwise sparsity & 0.05 & Evaluation episodes & 20\\
\bottomrule
\end{tabular}
\end{table}

CartPole has no physical no-op, so the neutral action of the intervention
library maps to the integer midpoint (the right action).  This asymmetry is a
limitation of using a common discrete library and is one reason not to
interpret the planner as an optimal controller.

\section{Development evidence and rejected variants}
\label{app:dev}

All choices below are excluded from the reported ten-seed analysis.
Architecture checks and the exact-copy diagnosis used seeds 0--69; a smoke gate
used seeds 70--72; the selected routed variant was then frozen; seeds 73--82
formed an earlier audit; and the reported ten-method comparison uses fresh
seeds 83--92.

\begin{table}[h]
\centering
\small
\caption{Development decisions.  These diagnostics are excluded from the
held-out result.}
\label{tab:development}
\begin{tabular}{lrrl}
\toprule
Variant & ID & Shift & Decision\\
\midrule
SOMBRL CartPole reference & 391.07 & 394.50 & smoke reference\\
Two-member CartPole route & 257.27 & 247.23 & reject\\
Three-member CartPole route & 473.40 & 462.40 & retain\\
Pendulum base controller & $-1278.28$ & $-1234.66$ & retain exactly\\
Low-scale causal planner & $-1278.28$ & $-1234.66$ & no gain\\
Stronger causal planner & $-1332.27$ & $-1268.54$ & reject\\
\bottomrule
\end{tabular}
\end{table}

Convex hybrids, a two-member route, and planner residuals were rejected.  The
agreement gate was retained because uncertain or contradictory advice then
cannot change the deterministic action.  Early qualification also considered
Acrobot and FrozenLake: under the fixed 4{,}096-interaction budget no
controlled method learned a useful Acrobot policy, while FrozenLake's discrete
grid and sparse reward do not match a continuous identified-coordinate
transition model, so reporting either would mostly measure benchmark mismatch.

\section{Statistical analysis details}

The seed is the unit of analysis.  For each benchmark and condition let
$y_{s,m}$ be the 20-episode mean for seed $s$ and method $m$; the descriptive
SD is $s_m=\sqrt{\frac{1}{9}\sum_{s=1}^{10}(y_{s,m}-\bar y_m)^2}$.  For the
matched comparison we compute $d_s=y_{s,\mathrm{CN}}-y_{s,\mathrm{route}}$ and
the two-sided interval $\bar d\pm t_{0.975,9}s_d/\sqrt{10}$.  Correlations in
\S\ref{sec:fidelity} are Pearson over the ten seeds; the Spearman analogues
agree in sign and significance ($F_1$ vs.\ $\Delta$ ID: $\rho_S=-0.23$,
$p=0.53$; NRMSE vs.\ $\Delta$ ID: $\rho_S=+0.59$, $p=0.07$).  Four cells and
several method comparisons make uncorrected significance hunting misleading, so
we emphasize effect sizes, intervals, per-seed differences, and average rank.
No bootstrap over evaluation episodes is used, because episodes within a
trained seed do not replace independent training replicates.

\section{Per-seed results}

\begin{figure}[h]
\centering
\includegraphics[width=0.78\linewidth]{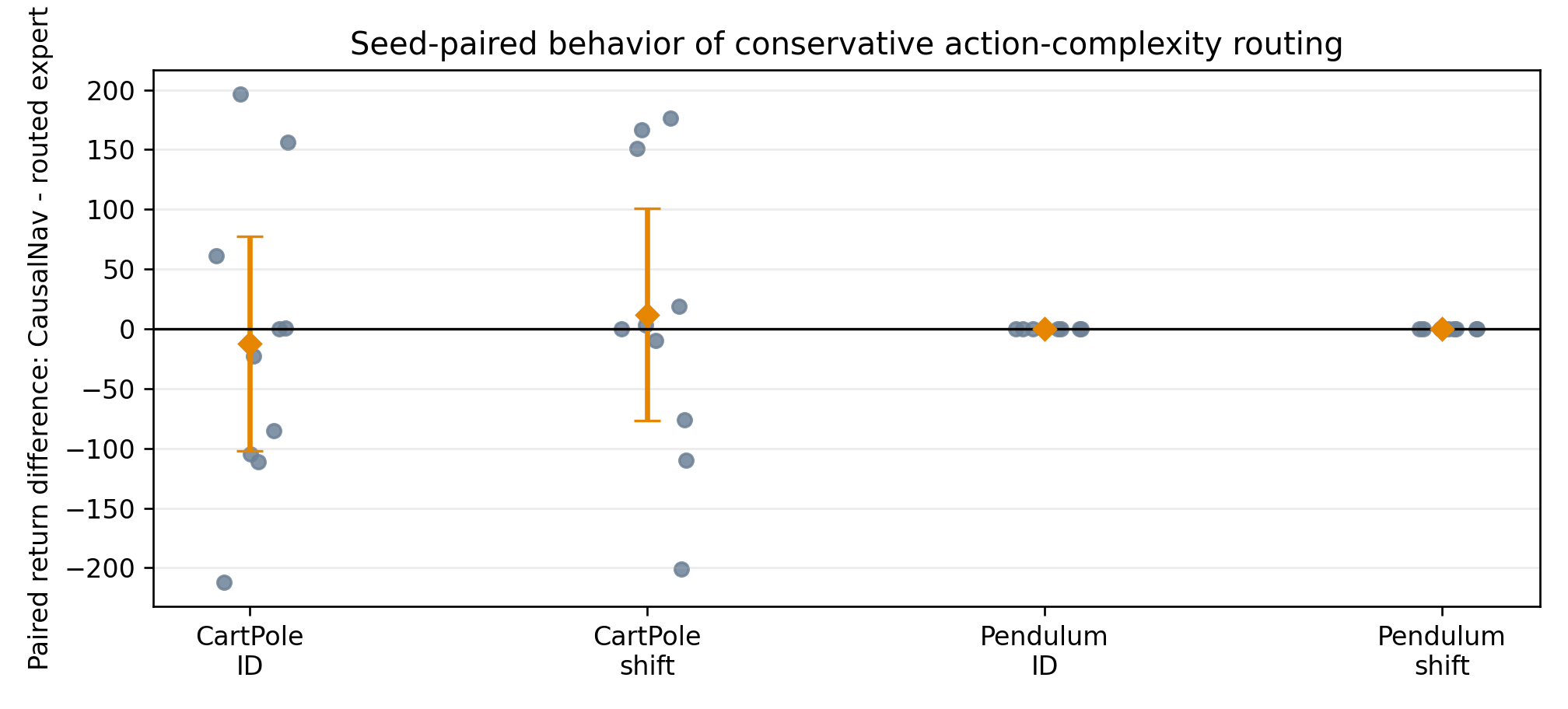}
\caption{Per-seed return difference between \method{} and its routed expert
(SOMBRL for CartPole; its own base controller for Pendulum).  Orange diamonds
are paired means;
bars are Student-$t$ 95\% confidence intervals.}
\label{fig:paired}
\end{figure}

Tables~\ref{tab:allcartid}--\ref{tab:allpendshift} report every seed--method
mean behind Table~\ref{tab:main}, exposing the pronounced bimodality of
CartPole and showing that no failed training run was removed.  Column
abbreviations: CN (\method{}), TF (TF-Policy), GRU (GRU-World), Split
(Split-Latent), GNN (GNN-RAG), EZ2 (EfficientZero V2), SOM
(SOMBRL), Flat (FlatMBRL).

\begin{table}[h]
\centering\scriptsize
\caption{CartPole-v1 in-distribution return by held-out seed.}
\label{tab:allcartid}
\begin{tabular}{r*{10}{r}}
\toprule
Seed&CN&TF&GRU&Split&GNN&ToG&CAASL&EZ2&SOM&Flat\\\midrule
83&496.9&20.3&146.2&35.5&476.8&159.9&336.8&187.3&496.6&421.0\\
84&359.9&414.1&228.7&238.9&500.0&337.3&250.6&137.9&464.6&140.2\\
85&500.0&384.6&477.8&175.1&499.5&500.0&450.6&181.7&343.7&486.6\\
86&500.0&140.2&500.0&481.1&207.6&500.0&287.2&274.5&438.8&199.8\\
87&388.8&468.8&184.6&495.6&449.6&259.8&185.8&243.3&500.0&210.1\\
88&300.1&494.9&203.4&426.0&130.9&144.3&172.2&270.9&103.6&449.9\\
89&169.2&351.1&381.2&500.0&144.2&500.0&101.7&254.2&254.7&205.4\\
90&288.1&11.4&326.6&183.1&144.2&182.3&349.4&264.6&500.0&201.9\\
91&500.0&351.1&281.4&248.2&227.9&405.5&500.0&500.0&500.0&150.4\\
92&219.6&9.8&170.8&37.0&119.8&85.0&239.9&328.0&242.1&263.1\\
\bottomrule
\end{tabular}
\end{table}

\begin{table}[h]
\centering\scriptsize
\caption{CartPole-v1 physical-shift return by held-out seed.}
\label{tab:allcartshift}
\begin{tabular}{r*{10}{r}}
\toprule
Seed&CN&TF&GRU&Split&GNN&ToG&CAASL&EZ2&SOM&Flat\\\midrule
83&500.0&24.8&150.8&45.1&446.4&158.6&278.6&179.7&481.4&400.1\\
84&314.6&343.4&242.8&211.2&500.0&326.5&231.8&167.1&311.2&195.2\\
85&500.0&388.9&478.7&178.1&495.0&500.0&443.9&175.2&333.4&500.0\\
86&500.0&142.5&500.0&480.9&208.6&500.0&289.9&266.6&323.4&167.7\\
87&390.1&436.9&194.7&488.9&447.1&260.8&186.4&242.2&500.0&212.2\\
88&259.0&492.4&207.8&396.1&121.2&150.7&176.0&272.2&108.2&460.6\\
89&176.2&330.2&354.4&500.0&148.8&500.0&105.8&248.7&252.1&209.2\\
90&298.8&17.4&308.7&187.9&147.3&181.4&350.8&259.6&500.0&199.5\\
91&500.0&343.6&268.4&251.4&220.2&363.9&500.0&500.0&500.0&152.6\\
92&226.3&12.2&189.8&53.7&112.2&88.9&237.8&330.6&235.8&288.7\\
\bottomrule
\end{tabular}
\end{table}

\begin{table}[h]
\centering\scriptsize
\caption{Pendulum-v1 in-distribution return by held-out seed.}
\label{tab:allpendid}
\begin{tabular}{r*{10}{r}}
\toprule
Seed&CN&TF&GRU&Split&GNN&ToG&CAASL&EZ2&SOM&Flat\\\midrule
83&-1259.5&-1285.0&-1463.1&-1259.5&-1245.6&-1361.8&-1414.5&-1144.2&-1316.1&-1392.9\\
84&-1303.9&-1124.9&-1100.5&-1114.3&-1172.8&-1285.2&-1145.0&-1410.1&-1391.2&-1269.0\\
85&-1406.0&-1441.3&-1382.3&-1551.0&-1372.4&-1381.5&-1395.8&-1296.3&-1272.0&-1215.2\\
86&-862.4&-1183.2&-1420.7&-1236.9&-1364.1&-1405.9&-1566.5&-1089.9&-1130.2&-1408.5\\
87&-1045.4&-1289.7&-1500.2&-1476.8&-1350.1&-1375.1&-1272.5&-1244.2&-1495.2&-1389.8\\
88&-1358.5&-1312.1&-1231.6&-1315.5&-1304.4&-1214.3&-1146.2&-1151.0&-1239.8&-1348.8\\
89&-1101.3&-1374.6&-1376.9&-1357.6&-1167.2&-1275.4&-1294.7&-1403.2&-1341.9&-1263.5\\
90&-1488.4&-1403.9&-1501.6&-1503.8&-1320.4&-1440.0&-1574.9&-1326.0&-1264.0&-1117.4\\
91&-1062.4&-1336.5&-1238.2&-1274.3&-1336.2&-1139.7&-1325.8&-1357.8&-1331.5&-1272.2\\
92&-1136.7&-1298.5&-1247.9&-1188.5&-1118.2&-1355.2&-1402.6&-1309.1&-1087.3&-1344.0\\
\bottomrule
\end{tabular}
\end{table}

\begin{table}[h]
\centering\scriptsize
\caption{Pendulum-v1 physical-shift return by held-out seed.}
\label{tab:allpendshift}
\begin{tabular}{r*{10}{r}}
\toprule
Seed&CN&TF&GRU&Split&GNN&ToG&CAASL&EZ2&SOM&Flat\\\midrule
83&-1331.2&-1312.3&-1329.2&-1343.3&-1323.8&-1387.2&-1454.6&-1262.4&-1361.6&-1410.8\\
84&-1159.3&-1073.8&-1101.9&-1200.2&-1193.7&-1213.7&-1227.7&-1192.4&-1166.6&-1255.9\\
85&-1363.3&-1346.4&-1330.2&-1472.0&-1336.7&-1327.2&-1294.3&-1265.0&-1257.2&-1244.7\\
86&-1054.2&-1160.8&-1362.6&-1158.7&-1336.9&-1437.7&-1497.3&-1098.0&-1162.2&-1382.7\\
87&-1139.6&-1236.9&-1391.6&-1342.5&-1303.3&-1318.3&-1220.3&-1161.0&-1302.3&-1331.8\\
88&-1228.2&-1289.2&-1198.2&-1289.5&-1251.2&-1210.6&-1104.0&-1107.4&-1159.5&-1279.3\\
89&-1157.1&-1289.7&-1295.6&-1313.9&-1233.6&-1187.5&-1293.0&-1347.7&-1213.1&-1271.9\\
90&-1395.7&-1341.7&-1272.7&-1300.7&-1280.4&-1312.4&-1306.9&-1282.0&-1186.6&-1174.6\\
91&-1195.0&-1302.7&-1129.4&-1230.9&-1322.2&-1237.3&-1299.7&-1304.7&-1230.6&-1196.7\\
92&-1163.8&-1230.3&-1115.2&-1163.2&-1096.9&-1244.5&-1317.5&-1256.8&-1201.5&-1312.6\\
\bottomrule
\end{tabular}
\end{table}

\section{Per-seed world-model diagnostics}

Table~\ref{tab:diag} gives the per-seed quantities behind
\S\ref{sec:fidelity}: structural $F_1$ against the CartPole ground-truth
adjacency, the exponential one-step NRMSE, the frozen certificate $\rho$, and
the paired return differences against SOMBRL.  Seeds with $\rho=0$ executed the
routed expert exactly.

\begin{table}[h]
\centering
\small
\caption{Per-seed \method{} diagnostics on CartPole-v1.}
\label{tab:diag}
\begin{tabular}{rrrrrr}
\toprule
Seed & Structural $F_1$ & One-step NRMSE & $\rho$ & $\Delta$ ID & $\Delta$ shift\\
\midrule
83 & 0.500 & 0.426 & 0.791 & $0.4$ & $18.6$\\
84 & 0.667 & 0.257 & 0.720 & $-104.8$ & $3.4$\\
85 & 0.714 & 1.427 & 0.000 & $156.3$ & $166.6$\\
86 & 0.500 & 0.532 & 0.000 & $61.2$ & $176.6$\\
87 & 0.545 & 0.678 & 0.031 & $-111.2$ & $-109.9$\\
88 & 0.571 & 0.860 & 0.232 & $196.5$ & $150.8$\\
89 & 0.545 & 0.149 & 1.000 & $-85.5$ & $-75.8$\\
90 & 0.727 & 0.282 & 0.136 & $-211.9$ & $-201.2$\\
91 & 0.667 & 0.323 & 0.339 & $0.0$ & $0.0$\\
92 & 0.500 & 0.655 & 0.000 & $-22.5$ & $-9.5$\\
\bottomrule
\end{tabular}
\end{table}

\section{Artifact map and implementation checks}

\begin{table}[h]
\centering
\small
\caption{Primary reproducibility artifacts.}
\label{tab:artifactmap}
\begin{tabular}{p{45mm}p{85mm}}
\toprule
Artifact & Purpose\\
\midrule
\texttt{agents/causalnav\_v2.py} &
Transition model, certificate, intervention planner, causal agent\\
\texttt{agents/baselines\_v2.py} &
Nine controlled reference architectures\\
\texttt{envs/*\_benchmark.py} &
Public benchmark adapters and structural metadata\\
\texttt{trainer.py} &
Shared PPO loop, paired initialization, successor targets, clipping\\
\texttt{run\_two\_benchmarks.py} &
Ten-seed training/evaluation driver with resumable JSON output\\
\texttt{assemble\_public\_results.py} &
Cardinality checks, summaries, paired intervals, frozen artifact\\
\texttt{plot\_results.py} &
All manuscript figures derived from the frozen artifact\\
\bottomrule
\end{tabular}
\end{table}

The released runner enforces the following invariants: (i) all methods receive
identical benchmark seeds, interaction budgets, PPO hyperparameters, and
identified-state targets; (ii) the active CartPole route uses three
deterministically seeded optimistic experts while the Pendulum route exactly
disables the causal component and executes the base controller unmodified;
(iii) optimizer state persists across rollouts;
(iv) auxiliary gradients reach the observation encoder; (v) transition targets
are actual successor states and never connect the final state of one episode to
the first state of another; and (vi) evaluation actions come from learned
logits without scripted outcome injection.

\end{document}